\documentclass[manuscript,screen]{acmart}
\usepackage{multirow}
\usepackage{subcaption}
\usepackage{xspace}
\usepackage[dvipsnames]{xcolor}
\definecolor{lightgray}{gray}{0.6}
\newcommand{\queryplot}{{\itshape QueryPlot}\xspace}
\newcommand{\hostrocks}{\textcolor{NavyBlue}{\texttt{host\_rocks}}\xspace}
\newcommand{\sourcerocks}{\textcolor{Maroon}{\texttt{source\_rocks}}\xspace}
\newcommand{\contact}{\textcolor{RoyalPurple}{\texttt{contact}}\xspace}
\AtBeginDocument{%
  }

\setcopyright{acmlicensed}
\copyrightyear{2025}
\acmYear{2025}
\acmDOI{XXXXXXX.XXXXXXX}
\acmConference[Conference acronym 'XX]{Make sure to enter the correct
  conference title from your rights confirmation email}{June 03--05,
  2018}{Woodstock, NY}
\acmBooktitle{ACM Trans. Spatial Algorithms Syst.}
\acmISBN{978-1-4503-XXXX-X/2018/06}

\begin{document}

\title{QueryPlot: Generating Geological Evidence Layers using Natural Language Queries for Mineral Exploration}

\author{Meng Ye}
\orcid{0009-0004-9290-2817}
\email{meng.ye@sri.com}
\affiliation{%
  \institution{SRI International}
  \city{Princeton}
  \state{NJ}
  \country{USA}
}

\author{Xiao Lin}
\orcid{0009-0003-0112-1699}
\email{xiao.lin@sri.com}
\affiliation{%
  \institution{SRI International}
  \city{Princeton}
  \state{NJ}
  \country{USA}
}

\author{Georgina Lukoczki}
\orcid{0000-0002-0661-0198}
\email{gina.lukozcki@uky.edu}
\affiliation{%
  \institution{University of Kentucky}
  \city{Lexington}
  \state{KY}
  \country{USA}
}

\author{Graham W. Lederer}
\orcid{0000-0002-9505-9923}
\email{grahamlederer@gmail.com}
\affiliation{%
  \institution{U.S. Geological Survey}
  \city{Reston}
  \state{VA}
  \country{USA}
}

\author{Yi Yao}
\orcid{0009-0009-0871-8758}
\email{yi.yao@sri.com}
\affiliation{%
  \institution{SRI International}
  \city{Princeton}
  \state{NJ}
  \country{USA}
}


\begin{abstract}
Mineral exploration is the process of identifying areas that are likely to host specific types of mineral deposits.
This task typically requires geologists to review extensive geological literature to understand the formation processes of different deposit types, extract common environmental attributes, and cross-reference them with geologic maps. This manual process is labor-intensive and can take years for a professional geologist to complete for a single deposit type.
In this work, we present \queryplot, a tool that leverages advanced Natural Language Processing (NLP) and Artificial Intelligence (AI) techniques to accelerate the mineral exploration workflow.
Specifically, we compile descriptive deposit model data for over 120 deposit types and process the State Geologic Map Compilation (SGMC) dataset to generate textual descriptions for each polygonal region. Given a user-defined query, our system computes semantic similarity scores between the query and each region using an embedding model and visualizes the results as an evidence layer for mineral potential mapping. Users can also query the database using various characteristics of a deposit type to generate multiple evidence layers, which can be accumulated together to derive prospective regions more effectively.
Through a case study on potential mapping of tungsten skarn deposits, we demonstrate, both qualitatively and quantitatively, that geological evidence layers generated by \queryplot can be used to retrieve known sites at a high recall rate, derive prospective regions with reasonable resemblance to permissive tracts generated by human experts, and serve as an additional input channel for training supervised machine learning models and increasing their classification performance.
\queryplot has a web-based interface that allows users to create queries, explore results interactively, and export prospectivity maps into file formats readable by other Geographic Information System (GIS) software for further geospatial analysis.
To support future research, we have made the source code and datasets used in this study publicly available.\footnote{\url{https://github.com/DARPA-CRITICALMAAS/sri-ta2-mappable-criteria}}
\end{abstract}



\received{06 June 2025}

\maketitle

\section{Introduction}
Critical minerals are essential for domains such as clean energy, digital infrastructure, and national security. For example, neodymium is used in high-performance magnets found in electric vehicle (EV) motors, wind turbines, and solar panel tracking systems. Lithium is a key component in lithium-ion batteries, which power portable electronics, EVs, and grid-scale energy storage systems. As industries move toward electrification and sustainable energy, the demand for these critical minerals is rapidly increasing.
For example, according to the International Energy Agency (IEA)~\cite{IEA_Critical_Minerals_2021}, demand for lithium is estimated to be growing by over 40 times in the sustainable development scenario, largely driven by its use in battery storage. Similarly, the demand for rare earth elements (REEs), e.g., neodymium, is set to more than triple by 2040.
This rising demand places significant pressure on global supply chains~\cite{USGS_Minerals_2023}, driving up costs and creating geopolitical challenges.
To increase the speed, efficiency, and sustainability of discovering and developing new critical mineral resources, geologists have increasingly turned to mineral prospectivity mapping (MPM) as a powerful tool for guiding exploration efforts. MPM is a method used to identify regions with a higher likelihood of hosting specific types of mineral deposits. By integrating various types of geoscientific data, including geological, geochemical, geophysical, and remote sensing datasets, geologists can analyze spatial patterns to prioritize areas for mineral exploration.
This substantially narrows down the areas requiring detailed site characterization, improving exploration efficiency and success rates.
However, the MPM process remains time-consuming and knowledge-intensive. It often involves reviewing historical exploration records, academic literature, and annotated geologic maps. Conducting an assessment manually for a single deposit type in complex terrains could take two or more years~\cite{CriticalMAAS_2023}, making it inefficient to evaluate all critical minerals in time to inform strategic planning for future resource availability.

\begin{figure}[t]
  \centering
  \includegraphics[width=\linewidth]{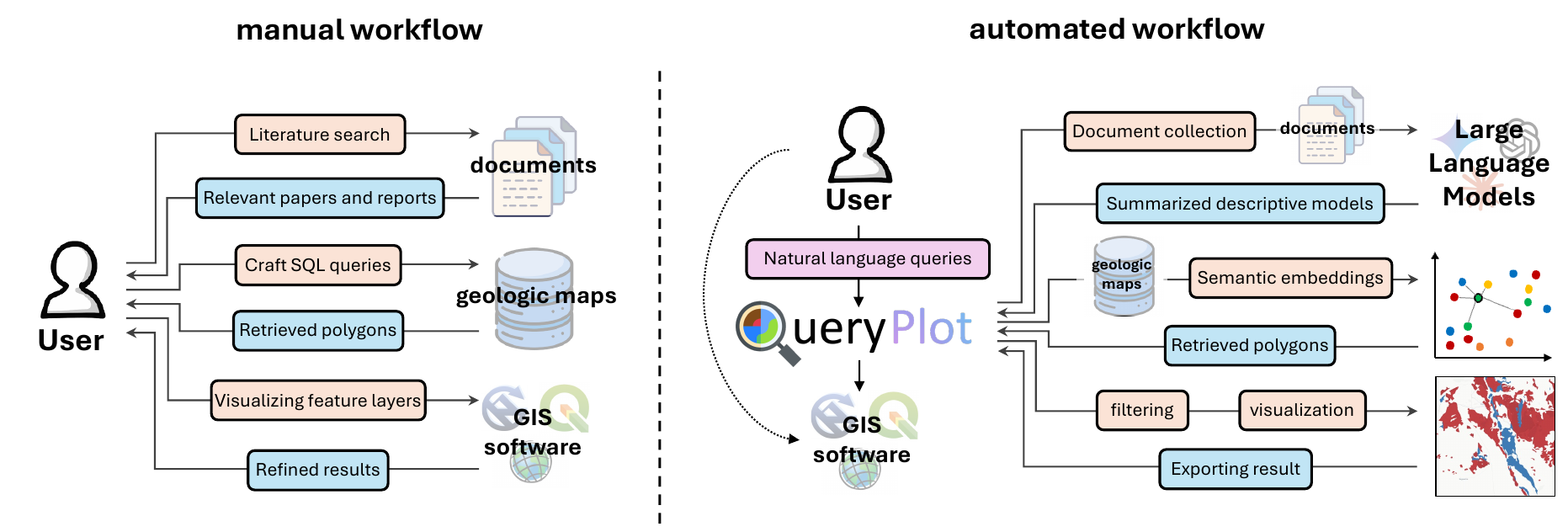}
  \caption{\queryplot helps geologists accelerate mineral exploration workflow through automating document summarization, geologic maps database retrieval, and evidence layer visualization. It offloads the burden of interacting with multiple tools and managing different data products from geologists.}
\label{fig:QueryPlot_workflow}
\end{figure}

To address these challenges, we developed \queryplot, an interactive tool designed for accelerating geologists' MPM workflow through automated generation of geological evidence layers from natural language queries.
For example, given a user input query in natural language, such as “dolostone and limestone in a carbonate platform sequence”, \queryplot embeds it into semantic feature space and compares this description with pre-ingested regional geological data. By calculating their similarity scores, high-ranking polygons are selected and visualized as a heatmap to indicate areas with geological evidence relevant to the query.
In addition to taking user-defined queries, \queryplot is also capable of extracting information to create descriptive deposit models automatically from relevant documents. These concise descriptive models contain characteristics that describe different aspects of a deposit type, guiding users to search for prospective regions quickly and effectively in a systematic manner.
\queryplot assists geologist users in several ways: First, it introduces a ``soft-scoring'' mechanism for querying geological databases using natural language, overcoming the limitations of rigid string-matching Structured Query Language (SQL) queries that struggle with diverse or ambiguous geological terminology. Second, it incorporates automated text summarization to condense lengthy documents such as geological reports and academic papers into concise deposit models, capturing the characteristic features (e.g., rock types, tectonic settings) of specific deposit types. Lastly, it offers a user-friendly web interface that integrates data ingestion, visualization, and export capabilities, reducing the need for switching between databases, GIS software, and other tools (Fig.~\ref{fig:QueryPlot_workflow}). To summarize, our contributions in this work are as follows:
\begin{itemize}
    \item We developed an NLP-based method to translate natural language geological queries into evidence layers for mineral exploration.
    \item We demonstrate the effectiveness of \queryplot through a case study on tungsten skarn deposit mapping in the southwestern United States.
    \item We implemented a prototype system that supports data loading, evidence map generation, interactive exploration, and exporting geospatial outputs.
\end{itemize}
The rest of this paper is organized as follows: section~\ref{sec:related_work} reviews related work and positions our approach in the context of existing research. Section~\ref{sec:methodology} describes the technical design of \queryplot and the use of transformer-based text encoders. Section~\ref{sec:evaluation} presents a case study evaluating the system on tungsten skarn deposits. Section~\ref{sec:prototype} outlines the user interaction flow through a web interface. Finally, section~\ref{sec:limitation} discusses limitations and directions for future research.

\section{Related work}
\label{sec:related_work}
\subsection{AI for geoscience}
There has been an increase in the application of Machine Learning (ML) techniques in geoscience studies. Earlier works include using Principle Component Analysis (PCA) and Support Vector Machines (SVM) to classify geomorphological features~\cite{mjolsness2001machine}, K-means for clustering lithogeochemical data in Zn skarn deposit exploration~\cite{jansson2022principal}, Logistic Regression and Random Markov Fields for spectral-spatial image segmentation~\cite{li2011spectral}, Random Forest for characterizing signature attributes of catchments~\cite{addor2018ranking}, etc.
With the recent rapid development of advanced AI technology, several works demonstrated AI's capability in handling big geoscience data~\cite{zhao2024artificial}. 
\cite{turarbek20232} uses Convolutional Neural Networks (CNN) for modeling seismic intensity.
\cite{kadow2020artificial} uses recently developed inpainting technique to reconstruct high-resolution global climate data on a monthly basis. \cite{khanna2023diffusionsat} trained a large generative foundation model on satellite imagery, enabling generation of realistic samples.
As Large Language Models (LLM) revolutionized the Natural Language Processing (NLP) field, LLM for geosciences also emerged~\cite{mai2024opportunities}.
K2~\cite{deng2024k2} was the first geoscience LLM. It was first pre-trained on geoscience text corpus and then instruction-tuned to align with human preference. The resulting GPT-like model can be used as a knowledgeable research assistant that can answer questions, provide ideas, and perform reasoning.
GeoGalactica~\cite{lin2023geogalactica} is a larger 30 billion parameter model trained on an expanded dataset from K2, capable of generating content, such as scientific papers, opinions, and paper summaries.
GeoGPT~\cite{zhang2024geogpt} is a framework developed by researchers for handling autonomous geospatial tasks, including data collection, processing, and analysis. It utilizes GPT4~\cite{hurst2024gpt}'s strong reasoning capability and externally defined GIS tools to comprehend complex queries, decompose user queries into subtasks, and execute subtasks step by step.
These works demonstrate that AI has been used to assist geology studies and holds huge potential in further accelerating geoscience advancement in the future.

\subsection{Mineral prospectivity mapping using computer vision models}
Mineral exploration is a sub-discipline of geoscience, focusing on finding and identifying mineral resources. MPM is a crucial tool in mineral exploration, acting as a guide to help identify areas with a higher likelihood of hosting undiscovered mineral deposits.
To accelerate the MPM process, researchers have been using computer vision models trained on known deposit sites for finding potential new deposits.
\cite{sun2020data} employed a set of machine learning methods—including random forest, SVM, artificial neural networks (ANN), and CNN—for data-driven tungsten prospectivity modeling in Jiangxi Province, China. Their approach revealed several novel exploration criteria via the modeling process, which could facilitate future tungsten prospecting.
Similarly, in~\cite{lou2023mineral}, the authors applied machine learning techniques to map tungsten polymetallic prospectivity in the Gannan region of China. By analyzing the contribution rates of sixteen evidential maps, they identified six elements closely related to tungsten polymetallic mineralization.
In~\cite{li2023mineral}, the authors proposed an attention-based CNN model that incorporates additional channel attention modules after each convolutional layer to enhance the extraction of key features relevant to mineralization. Their model demonstrated superior performance in prospectivity modeling for the Nanling metallogenic belt in southern China.
\cite{he2024mineral} utilized ensemble learning by training six different models (including VGG~\cite{simonyan2014very}, ResNet~\cite{he2016deep}, ViT~\cite{dosovitskiy2020image}, etc.) and combining their predictions for improved accuracy. More recently, \cite{daruna2024gfm4mpm} explored the use of self-supervised masked image modeling alongside supervised training frameworks to mitigate the data imbalance challenge frequently encountered in MPM.

All these studies adopt a similar supervised learning paradigm: computer vision models are trained to take evidence map layers as input and output prospectivity scores for each location in the map. The success of such approaches depends heavily on the availability and quality of input evidence layers, which provide the necessary information for the models to learn robust associations between geological features and mineralization potential. However, the selection and preparation of appropriate evidence map layers often require expert knowledge of the geological processes driving deposit formation. Furthermore, translating theoretical exploration criteria into numerical values for each region is a time-consuming and skill-intensive process. Therefore, automating the generation of evidence layers remains a key challenge for accelerating MPM workflows.


\subsection{Mineral exploration using Natural Language Processing}
Driven by advances in NLP technologies, researchers have begun leveraging language models for mineral prospectivity mapping tasks. For example, \cite{lawley2023applications} trained a word embedding model using rock descriptions, geological ages, and other long-form textual data from geological databases. Through network analysis of word associations, they demonstrated that the resulting embeddings captured meaningful semantic relationships between rock types. They further showed that these embeddings could be used to predict the distribution of certain rock types, estimate mineral potential (such as for Mississippi Valley-type (MVT) lead-zinc deposits), and search geoscientific text databases for analogues of deposits. In \cite{parsa2025pan}, the authors utilized word embeddings to generate evidence layers guided by conceptual exploration targeting criteria. Subsequently, a CNN with spatial attention mechanisms was trained to generate mineral deposit locations from input evidence maps. This pipeline was applied to predictive modeling of lithium-cesium-tantalum (LCT) pegmatites and successfully identified low-risk, high-return target areas in Canada.

These studies demonstrate the use of NLP models to analyze semantic information from large geological text corpora, applying this knowledge either to the generation of evidence layers or directly to the production of mineral prospectivity maps. Such approaches alleviate the burden of manual evidence layer preparation and, in some instances, may bypass the need for supervised training with labeled mineral occurrence sites. Our work is similar in that we also leverage semantic similarity scores to generate geological evidence layers. However, we extend prior work by introducing automated text summarization for the generation of descriptive deposit models, further simplifying the workflow. We also show that our generated layers can serve both as evidence for supervised predictive modeling and as effective stand-alone approximations of potentially permissive geologic units at the surface.

\section{Methodology}
\label{sec:methodology}
\subsection{System overview}

\begin{figure}[t]
  \centering
  \includegraphics[width=\linewidth]{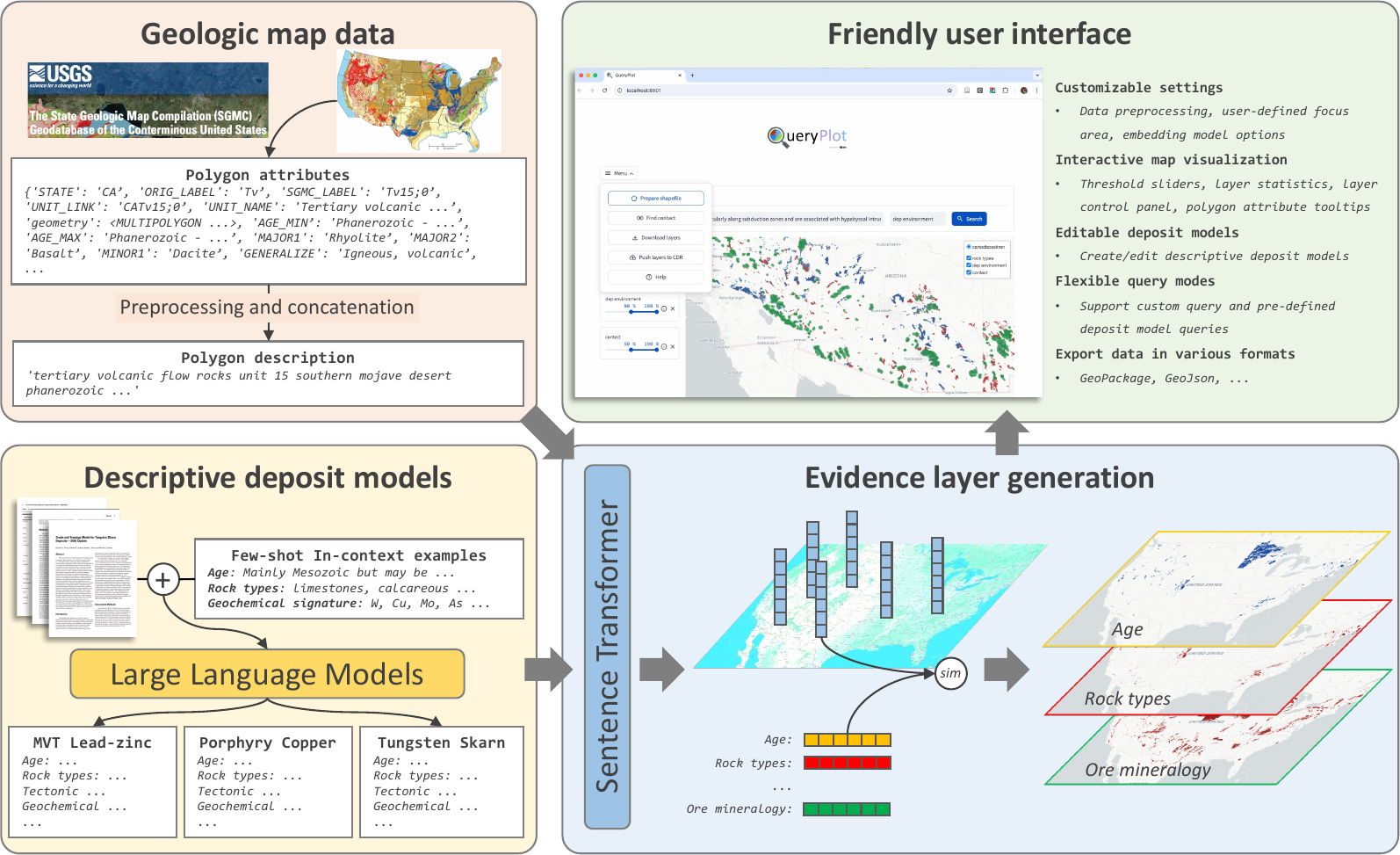}
  \caption{System overview of \queryplot. Upper left: Geologic map data are processed to generate long description for each polygon. Lower left: Long documents are summarized into descriptive deposit models using LLM. Lower right: Semantic similarity between textual queries and polygon descriptions are computed and visualized as evidence layers. Upper right: A prototype system with user-friendly UI helps geologists accelerate the mineral exploration workflow.}
\label{fig:overview}
\end{figure}

Our system comprises four key components: (1) geologic map data, (2) descriptive deposit model data, (3) evidence layer generation, and (4) a user interface. An overview of the system is provided in Fig.~\ref{fig:overview}.
We first collect geologic map data from authoritative sources such as SGMC. These data include geological polygons, associated tabular attributes, and, in some cases, unstructured textual descriptions. We preprocess these data by concatenating relevant attributes into a unified, extended description for each polygon.
Next, we collect academic papers and reports that study the formation processes and characteristics of various deposit types. We extract textual content from these documents and use LLMs to distill the key information into concise, structured descriptive deposit models. These models serve as the criteria for mapping regions of potential mineralization.
We then utilize transformer-based encoder models to extract sentence embeddings from both the geological polygon descriptions and the deposit models. The semantic similarity between each polygon and the user-defined query (or deposit model description) is computed and visualized as an evidence map specific to the query.
To facilitate user interactions, we developed an interface that integrates all system components, supporting configurable geologic map preprocessing, selection of embedding models, definition of focus areas, and interactive thresholding for output map filtering.
In the remainder of section~\ref{sec:methodology}, we detail the workflow for processing geologic map data, generating descriptive deposit models, and computing evidence layers. The implementation of the user interface and its key features are discussed in section~\ref{sec:prototype}.

\subsection{Geologic map data}
\begin{figure}[t]
  \centering
  \includegraphics[width=0.85\linewidth]{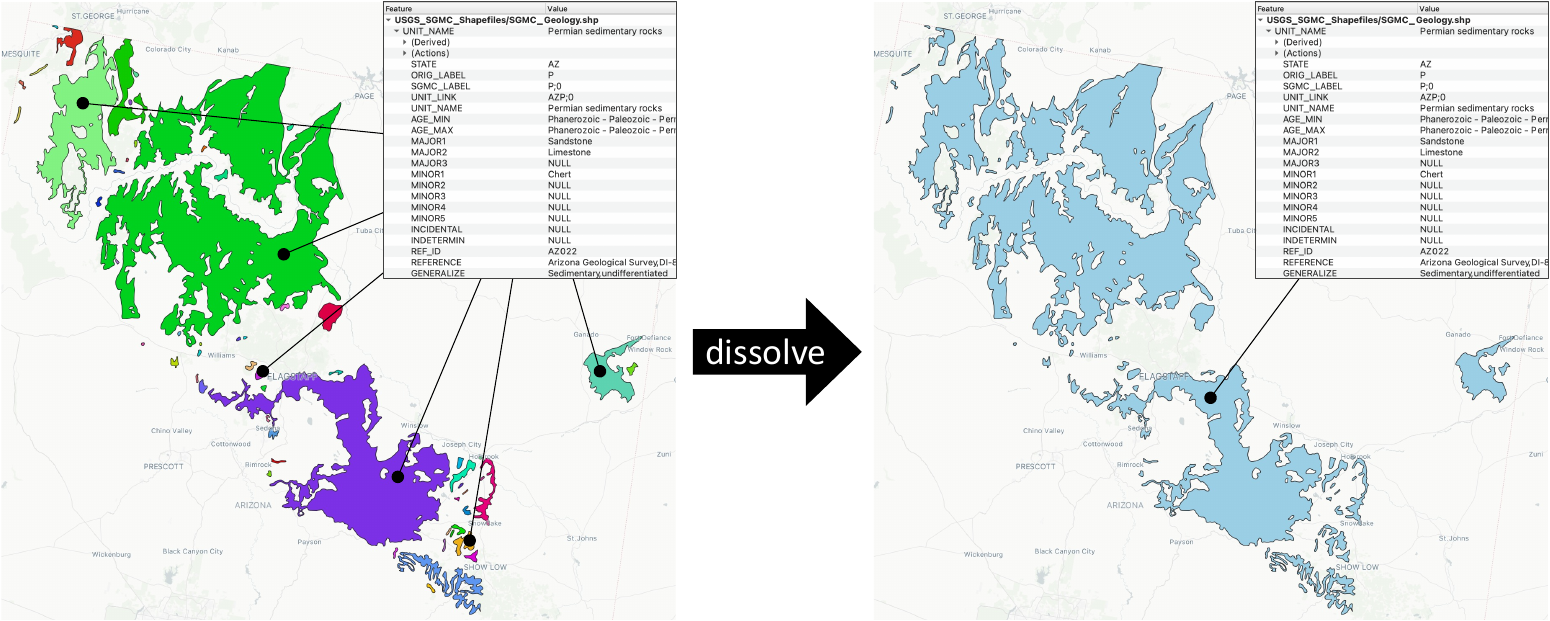}
  \caption{Merging multiple individual polygons with same geologic signatures into one multi-polygon.}
\label{fig:dissolve}
\end{figure}

We use SGMC~\cite{horton2017state} as our primary geologic map database. SGMC is a spatial database of geologic maps covering the conterminous United States, with map scales ranging from 1:50,000 to 1:1,000,000. It contains three feature classes (\texttt{SGMC\_Geology}, \texttt{SGMC\_Structure}, \texttt{SGMC\_Points}) and five tables (\texttt{Age}, \texttt{Lithology}, \texttt{Units}, \texttt{Unit\_Ref\_Link}, \texttt{References}). All data tables are provided in formats including ESRI shapefile and comma-separated values (CSV).
To facilitate effective querying for mineral prospectivity, we manually selected a subset of eleven attributes as the "signature" for each polygon. These include: The name of the map unit as given on the source map (\texttt{UNIT\_NAME}), three lithology attributes that were determined as ``major'' ranking (\texttt{MAJOR1-MAJOR3}), five lithology attributes that were dtermined as ``minor'' ranking (\texttt{MINOR1-MINOR5}), generalized rock classifications of the unit using the \texttt{MAJOR1} through \texttt{MAJOR3} fields (\texttt{GENERALIZE}), and the unit description as given on the source paper map (\texttt{UNITDESC}). These attributes are concatenated to form a comprehensive textual description (\texttt{full\_desc}) for each record.
We further process these descriptions with several cleaning steps, such as removing extra spaces and newlines, converting or removing non-ASCII characters, and filtering out records with insufficiently detailed descriptions. After this cleaning, the description column is prepared for embedding extraction, which will be described in section~\ref{sec:embedding}.

An important consideration is that the raw SGMC shapefile contains over 300,000 rows, with each row representing a spatial polygon and its associated attributes. 
These attributes include ``key'' columns that can be used to index each individual unit, such as two-letter state abbreviation (\texttt{STATE}), map unit label from the original source (\texttt{ORIG\_LABEL}), error-corrected and coded map unit symbol (\texttt{SGMC\_LABEL}), and a unique value generated by combining multiple other fields (\texttt{UNIT\_LINK}). Many records with identical key columns also have the same attributes and descriptions while only differ in spatial location. Repeatedly computing semantic similarity for these duplicate descriptions introduces unnecessary computational overhead. To address this, we deduplicated rows with matching key columns and merged the corresponding polygons, as shown in Fig.~\ref{fig:dissolve}. This "dissolve" operation reduced the dataset from approximately 300,000 polygons to about 7,000 multi-polygons, significantly speeding up the semantic search while preserving result accuracy.
We uses this "disolved" version of SGMC database for all the experiments in this paper. Nevertheless, \queryplot can be easily adapted to other geological databases that contain similar tabular data to SGMC.

\subsection{Descriptive deposit models}
\begin{table}[t]
\caption{An example of selected characteristics of the descriptive model for tungsten skarn deposits from~\cite{day2016overview}.}
\label{tab:dep_model}
\small
  \begin{tabular}{p{0.8in}p{4.5in}}
    \toprule
    {\bfseries Characteristic} & {\bfseries Description} \\
    \midrule
    \texttt{Synonyms} & \texttt{Pyrometasomatic or contact metasomatic tungsten deposits.} \\
    \hline
    \texttt{Commodities}& \texttt{W, Mo, Cu, Sn, Zn} \\
    \hline
    \texttt{Description}& \texttt{Scheelite-dominant mineralization genetically associated with a skarn gangue.}\\
    \hline
    \texttt{Rock types}& \texttt{Pure and impure limestones, calcareous to carbonaceous pelites. Associated with tonalite, granodiorite, quartz monzonite and granite of both I- and S-types. W skarn-related granitoids, compared to Cu skarn- related plutonic rocks, tend to be more differentiated, more contaminated with sedimentary material, and have crystallized at a deeper structural level.} \\
    \hline
    \texttt{Textures}& \texttt{Porphyry has closely spaced phenocrysts and microaplitic quartz-feldspar groundmass.} \\
    \hline
    \texttt{Age range}& \texttt{Mainly Mesozoic, but may be any age.} \\
    \hline
    \texttt{Depositional environment}& \texttt{Contacts and roof pendants of batholith and thermal aureoles of apical zones of stocks that intrude carbonate rocks.} \\
    \hline
    \texttt{Tectonic setting}& \texttt{Continental margin, synorogenic plutonism intruding deeply buried sequences of eugeoclinal carbonate-shale sedimentary rocks. Can develop in tectonically thickened packages in back-arc thrust settings.} \\
    \hline
    \texttt{Alteration}& \texttt{Exoskarn alteration: Inner zone of diopside-hedenbergite $\pm$ grossular-andradite $\pm$ biotite±vesuvianite, with outer barren wollastonite-bearing zone. An innermost zone of massive quartz may be present. Late stage spessartine $\pm$ almandine±biotite $\pm$ amphibole $\pm$ plagioclase $\pm$ phlogopite $\pm$ epidote $\pm$ fluorite±sphene. Exoskarn envelope can be associated with extensive areas of biotite hornfels. Endoskarn alteration: Pyroxene±garnet $\pm$ biotite $\pm$ epidote $\pm$ amphibole $\pm$ muscovite $\pm$ plagioclase $\pm$ pyrite $\pm$ pyrrhotite $\pm$ trace tourmaline and scapolite; local greisen developed.} \\
    \hline
    \texttt{Ore controls}& \texttt{Carbonate rocks in extensive thermal aureoles of intrusions; gently inclined bedding and intrusive contacts; structural and (or) stratigraphic traps in sedimentary rocks and irregular parts of the pluton/country rock contacts.} \\
    
    \bottomrule
  \end{tabular}
\end{table}

A mineral deposit model is the systematically arranged information describing the essential attributes (characteristics) of a class of mineral deposits, or, a deposit type~\cite{cox1986mineral}. 
Such models integrate geologic, geophysical, geochemical, and mineralogical data to help geologists understand where minerals are likely to occur.
One of the earliest and most notable compilations is the Cox \& Singer models~\cite{cox1986mineral}, which introduced the concept of descriptive deposit models. They have two components: (1) the ``Geological environment'', which characterizes the setting in which deposits are found, and (2) the ``Deposit description'', which details the identifying characteristics of the deposits, with an emphasis on geochemical and geophysical aspects.
It is important to note that a deposit model always corresponds to a specific deposit type—that is, a group of mineral deposits sharing similar geological characteristics and origin.
The classification scheme for deposit types has evolved over time. In this paper, we follow the Critical Minerals Mapping Initiative (CMMI) classification~\cite{hofstra2021deposit} for critical mineral mapping, which comprises 189 deposit types classified by mineral system.
It is also worth mentioning that, besides descriptive deposit models, there are also genetic deposit models, which focus on the origin and formation processes of mineral deposits.
Our study focuses on descriptive deposit models, however, the same methodology could be applied to genetic models, as these are also expressed in natural language.

\subsubsection{Document collection and parsing}
\begin{figure}[t]
  \centering
  \includegraphics[width=0.75\linewidth]{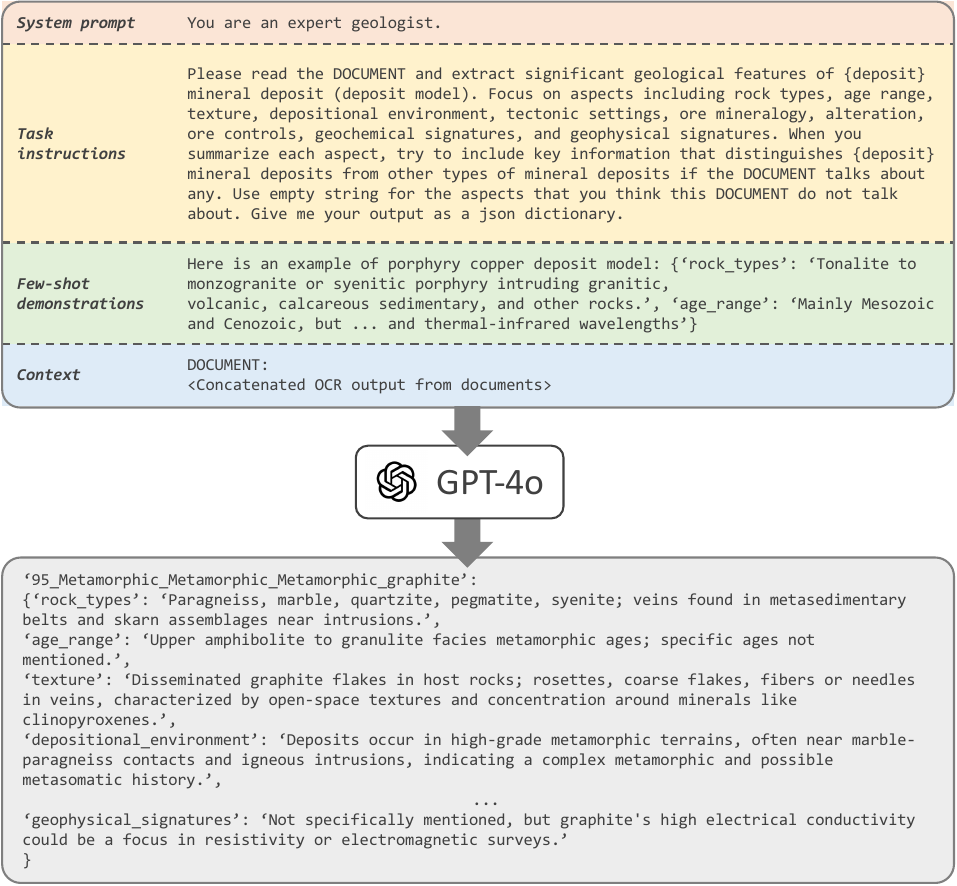}
  \caption{Example prompt for summarizing documents into short and concise descriptive deposit models.}
\label{fig:prompt}
\end{figure}
We collected 151 documents of deposit model studies from multiple sources, including those from Cox \& Singer~\cite{cox1986mineral}, reports from multiple geological survey agencies, and academic papers.
These documents cover 120 deposit types defined in the CMMI classification scheme~\cite{hofstra2021deposit}.
One example of a deposit model of the `tungsten skarn' deposit type acquired from~\cite{day2016overview} is shown in Table~\ref{tab:dep_model}.
We parsed the pdf documents and extracted their text content using the Tesseract OCR engine\footnote{\url{https://github.com/tesseract-ocr/tesseract}}.
For deposit types with multiple documents, we concatenated the text into a single, comprehensive document for each type.
These documents provide essential information about the characteristics of each deposit type, although this information is often not explicitly summarized or presented in a structured format.

\subsubsection{Summarizing long documents using LLM}

Long documents present challenges for both embedding extraction and semantic similarity computation, as typical embedding models (e.g., bge-base-v1.5~\cite{xiao2024c}) support a context window of only up to $512$ tokens (approximately $400$ words).
Although models capable of processing up to $32,768$ tokens have been proposed~\cite{zhu2024longembed}, they still suffer from the `Length Collapse' phenomenon~\cite{zhou2024length}, in which embeddings of long texts tend to cluster in a narrow region of feature space, leading to degraded performance in semantic retrieval tasks.

To obtain uniform, structured, and concise descriptions, we use LLMs to summarize all long documents into descriptive models comparable in length and structure to those in Cox \& Singer~\cite{cox1986mineral}.
LLMs have demonstrated outstanding performance in a variety of NLP tasks, including summarization~\cite{goyal2022news, liu2022revisiting, zhang2024benchmarking}, question answering~\cite{brown2020language, chowdhery2023palm}, and others.
Despite this success, LLMs are still subject to challenges such as loss of important details, hallucination, and focus drift~\cite{ramprasad2024analyzing, huang2025survey, laban2023summedits}.
To improve the reliability of the generated deposit models, we provide the entire concatenated document as input to the LLM, ensuring full context coverage. We also incorporate few-shot in-context demonstrations to improve the model's ability to follow instructions and produce outputs in the desired format. An example prompt is shown in Fig.~\ref{fig:prompt}.
For all the $120$ deposit types, we used GPT-4o\footnote{OpenAI API with model name \texttt{`gpt-4o-2024-08-06'}}~\cite{hurst2024gpt} to generate concise descriptive deposit models from long documents.


\subsection{Evidence layer generation}
\label{sec:embedding}
With the polygon descriptions and descriptive deposit models prepared, we can now generate evidence layers for any specified deposit type by comparing each characteristic in the deposit model with all polygons in the geologic map database.
Specifically, Let $\mathcal{P} = \{P_1, P_2, \ldots, P_N\}$ represent a set of $N$ polygons. Each polygon $P_i$ is associated with:
\begin{itemize}
    \item an ordered list of points defining exterior boundary of the polygon $\mathbf{C}_i = [(x_i^{(1)}, y_i^{(1)}), (x_i^{(2)}, y_i^{(2)}), \ldots, (x_i^{{n_i}}, y_i^{(n_i)})]$, where $x_i\in\mathbb{R}$ and $y_i\in\mathbb{R}$ denotes the latitude/longitude coordinates of each point.
    \item a set of attributes $A_i=\{(h_i^{(1)}, d_i^{(1)}), (h_i^{(2)}, d_i^{(2)}), \ldots, (h_i^{(m_i)}, d_i^{(m_i)})\}$, where $h_i\in \mathbb{T}$ denotes the headings (characteristic names), and $d_i\in \mathbb{T}$ denotes the textual description for each characteristic.
\end{itemize}

Let $\mathcal{M}_\theta: \mathbb{T} \rightarrow \mathbb{R}^d$ be a transformer~\cite{vaswani2017attention}-based sentence embedding function parameterized by weights $\theta$, mapping a textual string to a $d$-dimensional embedding vector.

Given a user query $q \in \mathbb{T}$, we compute its embedding as:
\begin{equation}
\mathbf{q} = \mathcal{M}_\theta(q)
\end{equation}

For each polygon $P_i$, we compute the embedding of its description:
\begin{equation}
\mathbf{v}_i = \mathcal{M}_\theta\left(\text{concat}([d_i^{(1)}, d_i^{(2)}, \ldots, d_i^{(m_i)}])\right)
\end{equation}

We then calculate the cosine similarity between the query embedding and each polygon embedding:
\begin{equation}
s_i = \cos(\mathbf{q}, \mathbf{v}_i) = \frac{\mathbf{q} \cdot \mathbf{v}_i}{\|\mathbf{q}\| \cdot \|\mathbf{v}_i\|}
\end{equation}

Let $\tau \in (0, 1]$ be a percentile threshold. We define the selected set of polygons as:
\begin{equation}
\mathcal{P}_\tau = \left\{P_i \in \mathcal{P} \mid s_i \text{ is among the top } \tau \cdot N \text{ similarity scores} \right\}
\end{equation}

The resulting evidence layer, which consists of top-ranked polygons, is constructed by merging the coordinates of the selected polygons:
\begin{equation}
\mathcal{E}_q = \bigcup_{P_i \in \mathcal{P}_\tau} \mathbf{C}_i
\end{equation}

\begin{figure}[t]
  \centering
  \includegraphics[width=\linewidth]{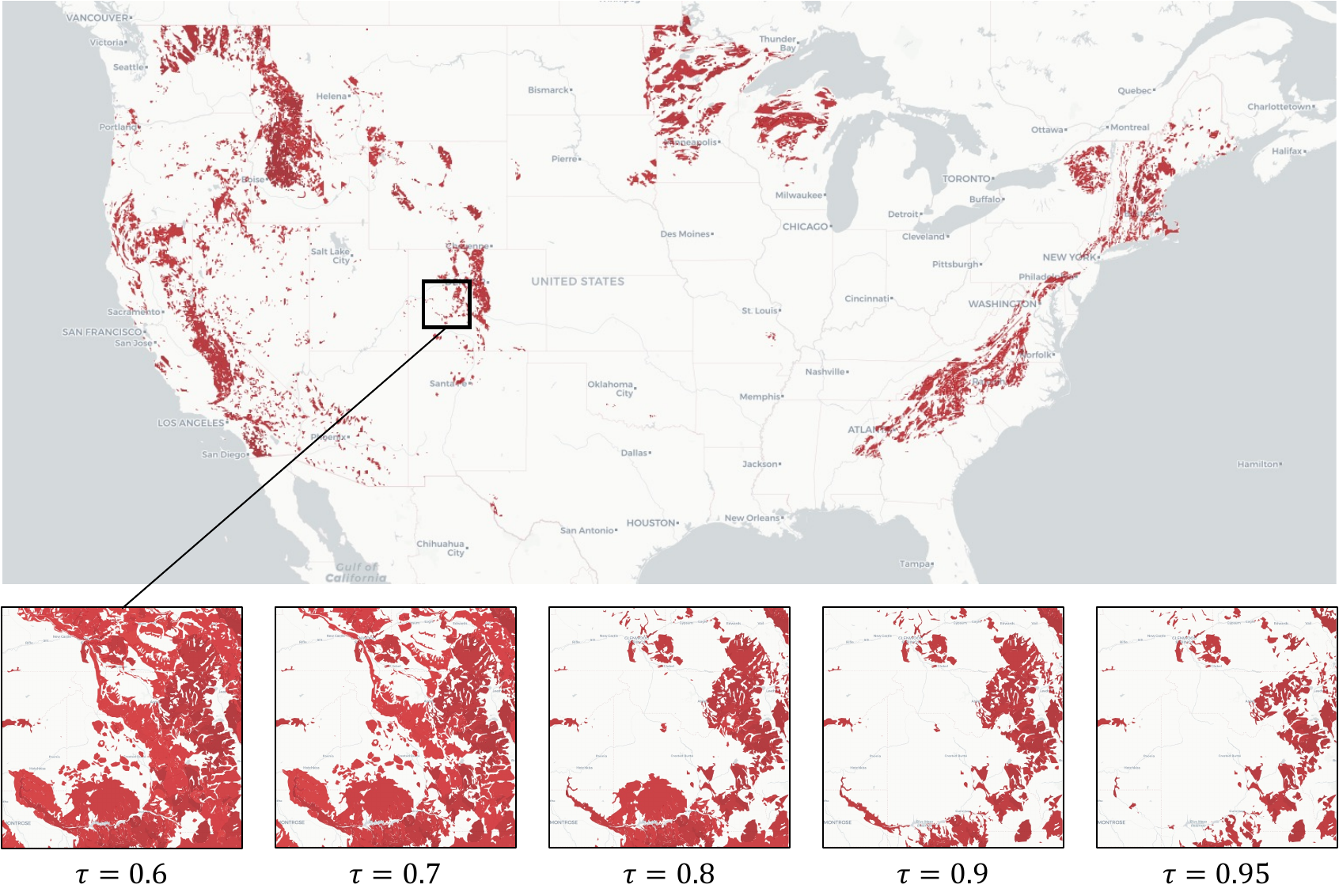}
  \caption{An evidence layer generated for national scale tungsten skarn exploration. The layer was generated from query \texttt{``tonalite, granodiorite, quartz monzonite and granite.''}, which includes the most common source rocks characteristic of tungsten skarn deposit model. Zoomed-in regions at the bottom shows the effect of different percentage threshold values, with higher threshold resulting in fewer polygons.}
\label{fig:thresholding}
\end{figure}

An example evidence layer generated for nation-wide tungsten skarn potential mapping is provided in Fig.~\ref{fig:thresholding}, illustrating the effects of varying the threshold parameter $\tau$.

To explore potential mineral deposit regions more effectively, multiple evidence layers $\mathcal{E}_Q=\{\mathcal{E}_{q_i}\}_{i=1}^l$ can be generated simultaneously based on a set of queries $Q=\{q_1,q_2,\ldots,q_l\}$, each representing a different deposit model characteristic.
Regions where multiple characteristics yield high similarity scores are thus considered more prospective.
For example, tungsten skarn deposits are associated with the {\bfseries contact zone} between host rocks (e.g., limestones) and source rocks (e.g., granites)~\cite{hammarstromsn, eckstrand1984canadian, green2020grade}. 
To identify regions with high mineral prospectivity, we consider polygons with high similarities to both source and host queries.
Additionally, because geologic bodies are typically mapped as adjacent—rather than overlapping—polygons, their shared edges usually represent surface-level contacts. However, in the case of skarn deposits, thermal interactions between the host rock and the source can extend beyond these contacts. Therefore, identifying overlapping areas between strong source and strong host polygons is also important.
To capture such spatial relationships, we apply buffer to each polygon, extending their boundaries outward. We then identify overlapping areas between buffered high-``source'' and high-``host'' polygons, effectively modeling spatial interactions between distinct but neighboring geologic units and highlighting zones where favorable geological features co-occur in close proximity.
More specifically, let $\text{Buffer}_r(\cdot)$ be the buffering operation, which creates a new polygon that expands outward from the boundaries of an existing polygon by a specific distance $r$. We buffer all polygons in all evidence layers:
\begin{equation}
\label{eq:buffer1}
    \mathcal{E}_q^\ast = \bigcup_{P_i\in\mathcal{E}_q}\text{Buffer}_{r_1}(P_i),\quad \text{for}\ \mathcal{E}_q\in\mathcal{E}_Q
\end{equation}
where $r_1$ is the buffer distance in the same coordinate reference system (CRS) (e.g., ESRI:102008) as the polygons. We then perform geometric intersection operation to all the buffered layers:
\begin{equation}
\label{eq:intersect}
    \mathcal{E}^+ = \text{Buffer}_{r_2}\left(((\mathcal{E}_{q_1}\cap\mathcal{E}_{q_2})\cap\mathcal{E}_{q_3})\cdots\cap\mathcal{E}_{q_l}\right)
\end{equation}
where $\cap$ denotes the operation of finding overlapping regions of two layers, which is extended from the intersection operation that identifies overlapping area of two individual polygons:
\begin{equation}
\label{eq:buffer2}
    \mathcal{E}_x\cap\mathcal{E}_y=\bigcup_{\text{for all}\ P_i\in\mathcal{E}_x,P_j\in\mathcal{E}_y}\mathbf{C}_i\cap\mathbf{C}_j
\end{equation}
\begin{figure}[t]
  \centering
  \includegraphics[width=0.95\linewidth]{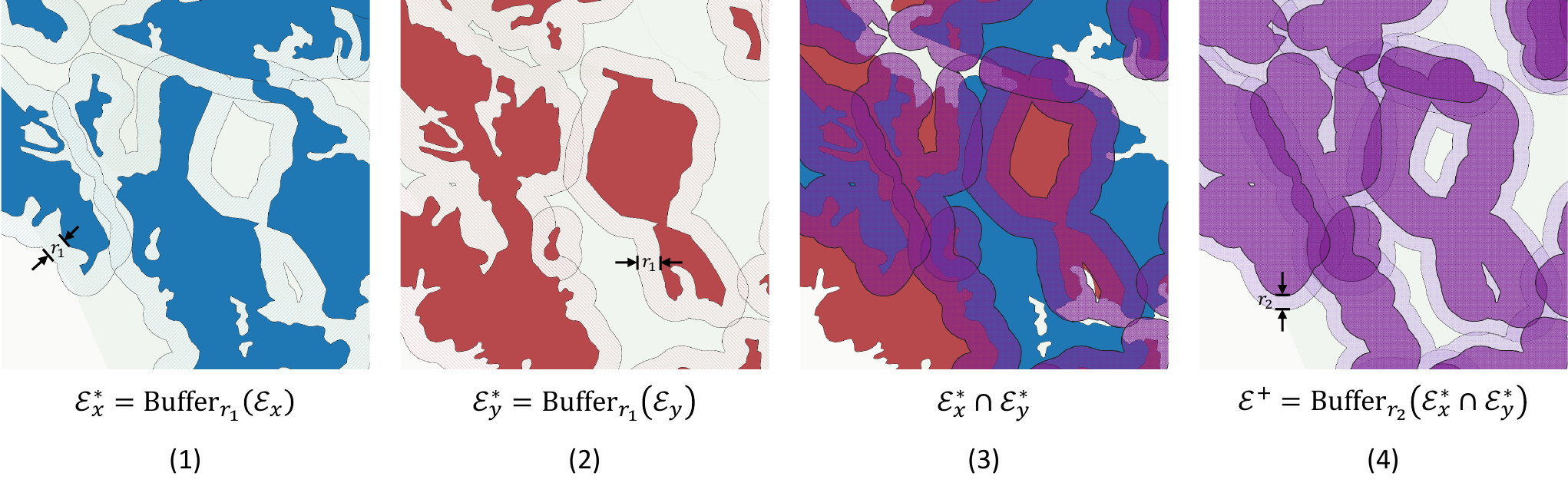}
  \caption{An example of finding high-potential areas adjacent to prospective regions from two evidence layers. (1) and (2) -- Buffer two evidence layers by a distance of $r_1$. (3) -- calculate the overlapping region of the buffered layers. (4) -- Buffer the intersected area by a distance of $r_2$.}
\label{fig:contact}
\end{figure}
Note that we performed one more buffering operation to the intersected layers with a different distance denoted as $r_2$. It is used for compensating narrow regions derived from intersection of polygons with short $r_1$ buffer distance. Practically, $r_1$ and $r_2$ are two hyperparameters that need to be tuned to reach desired output. Fig.~\ref{fig:contact} provides a visual example of this process on two evidence layers.

\section{Case study: Tungsten Skarn potential mapping in western United States}
\label{sec:evaluation}
\begin{figure}[t]
  \centering
  \includegraphics[width=0.95\linewidth]{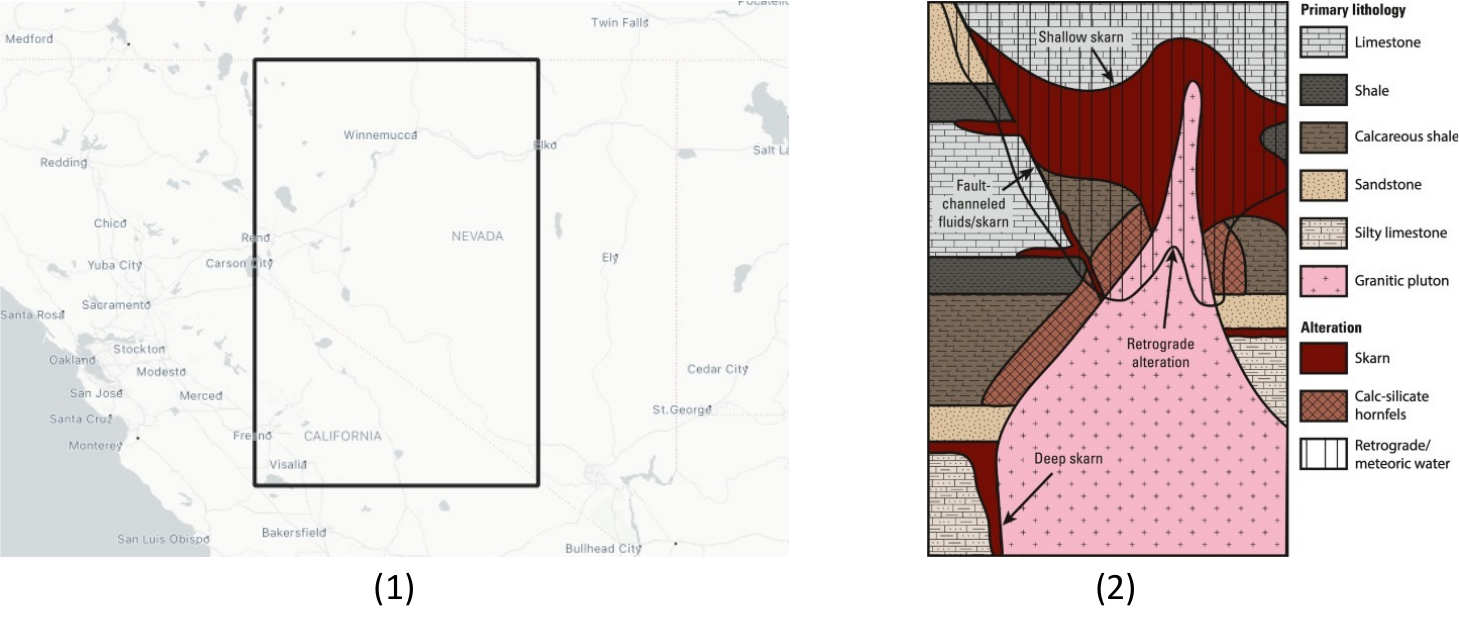}
  \caption{(1) The focus area of Tungsten skarn potential mapping case study. (2) Schematic representation of skarn deposits~\cite{lederer2021tungsten}}
\label{fig:focus_area}
\end{figure}
In this section, we demonstrate the application of \queryplot with a case study on Tungsten Skarn potential mapping in the western United States. We first define a study area which corresponds to that for a tungsten skarn assessment~\cite{lederer2021tungsten} of the Great Basin region. The area spans from $36^\circ$ to $42^\circ$ North latitude and from $120^\circ$ to $116^\circ$ West longitude, covering parts of eastern California and western Nevada.
Most skarn deposits form during the transfer of heat, fluid and metals from the cooling magma into the surrounding rock, through three stages. In the first stage, the magma intrudes and heats the bedrock without forming minerals. In the second stage, fluids escape from the magma and move through the surrounding rocks, initiating minor mineral deposition. The majority of ore minerals form in the third stage when the system cools, fluids mix with surface water, and reactions with carbonate occur. A schematic representation of this formation process is shown in Fig.~\ref{fig:focus_area}.
As mentioned earlier, tungsten skarn deposits are associated with the {\bfseries contact zone} between host and source rocks.
{\bfseries Source} rocks are typically relatively evolved felsic intrusive rocks, such as tonalite, granodiorite, quartz monzonite, and I- or S-type granites. {\bfseries Host} carbonate lithologies may include pure to impure limestones, dolostone, marble, and calcareous to carbonaceous pelites.
Thus, we base our queries on these studies and search for host rocks with \texttt{``limestones, calcareous to carbonaceous pelites.''}, and source rocks with \texttt{``tonalite, granodiorite, quartz monzonite and granite.''} as the descriptions.
We extract corresponding sentence embeddings for these two queries and perform semantic searches among SGMC polygons following the process described in section \ref{sec:methodology}.
\begin{figure}[t]
  \centering
  \includegraphics[width=0.95\linewidth]{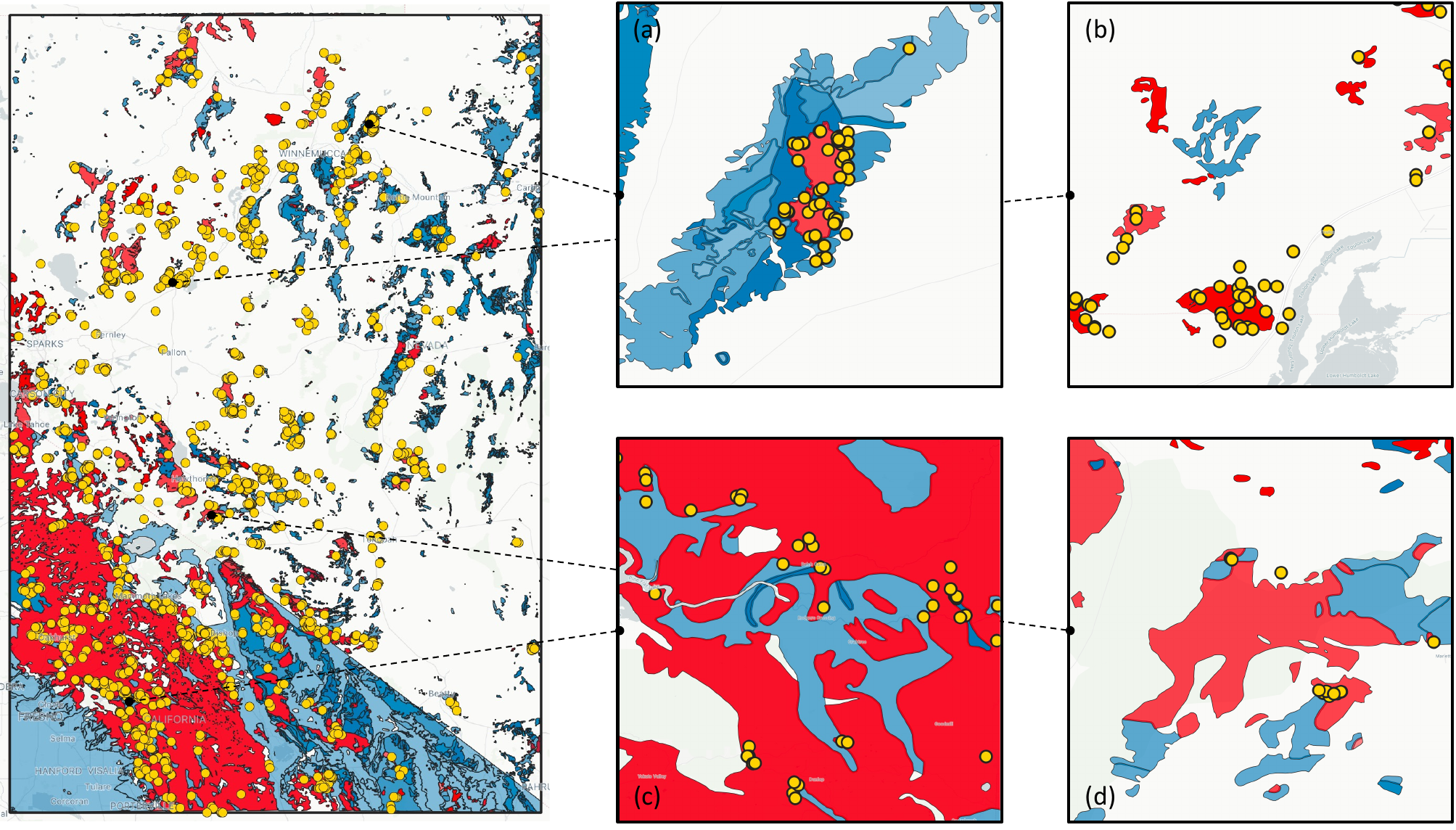}
  \caption{Comparison of generated evidence layers and mineral sites (blue regions: high-ranking polygons in the \hostrocks layer, red regions: high-ranking polygons in the \sourcerocks layer, yellow circles: mineral sites). (a)(c)(d): most sites are located near the contact regions of host and source rocks. (b): some sites do not lie on contact region but are captured by the source evidence layer.}
\label{fig:contact_known_sites}
\end{figure}

The resulting evidence layers (\hostrocks layer and \sourcerocks layer) are shown in Fig.~\ref{fig:contact_known_sites}.
Two evidence layers are overlaid together for easier comparison. High intensity blue and red regions denote stronger potential for host and source rocks, respectively. We have filtered out polygons that have low similarity scores to both queries, to focus on the top-ranked ones.
Yellow circle markers denote known mineral sites, including deposits, prospects, and occurrences.
We collected these sites from the Mineral Resources Data System (MRDS)\footnote{\url{https://mrdata.usgs.gov/mrds/}} and use them as a reference for qualitative check on the generated evidence layers.
We observe in the map that:
(1) Both evidence layers produce score distributions that differentiate the prospectivity of various polygons and regions.
(2) The spatial distribution of scores in the \hostrocks evidence layer differs from that in the \sourcerocks layer, as shown in the map that they highlight different regions. This demonstrates that the model is able to embed the two queries differently based on their semantic content. For example, the \sourcerocks layer indicates higher prospectivity (bright red) in the southwestern region of the study area than \hostrocks layer (light blue). This aligns well with geological studies in southern California, which have identified extensive exposures of granitic rocks indicating high potential for skarn-related mineralization.~\cite{ague1987granites}
(3) We found in many cases that known mineral sites lie close to the contact regions of host and source rocks (Fig.~\ref{fig:contact_known_sites}: (a), (c), and (d)), which aligns with the theoretical formation process of tungsten skarn.
Some sites do not have host rock types found near them (Fig.~\ref{fig:contact_known_sites}: (b)), but are still captured by the source rock types evidence layer, which can be explained by limitation of map scale that does not resolve small bodies of permissive host rocks.

\begin{figure}[t]
    \centering
    \includegraphics[width=0.5\linewidth]{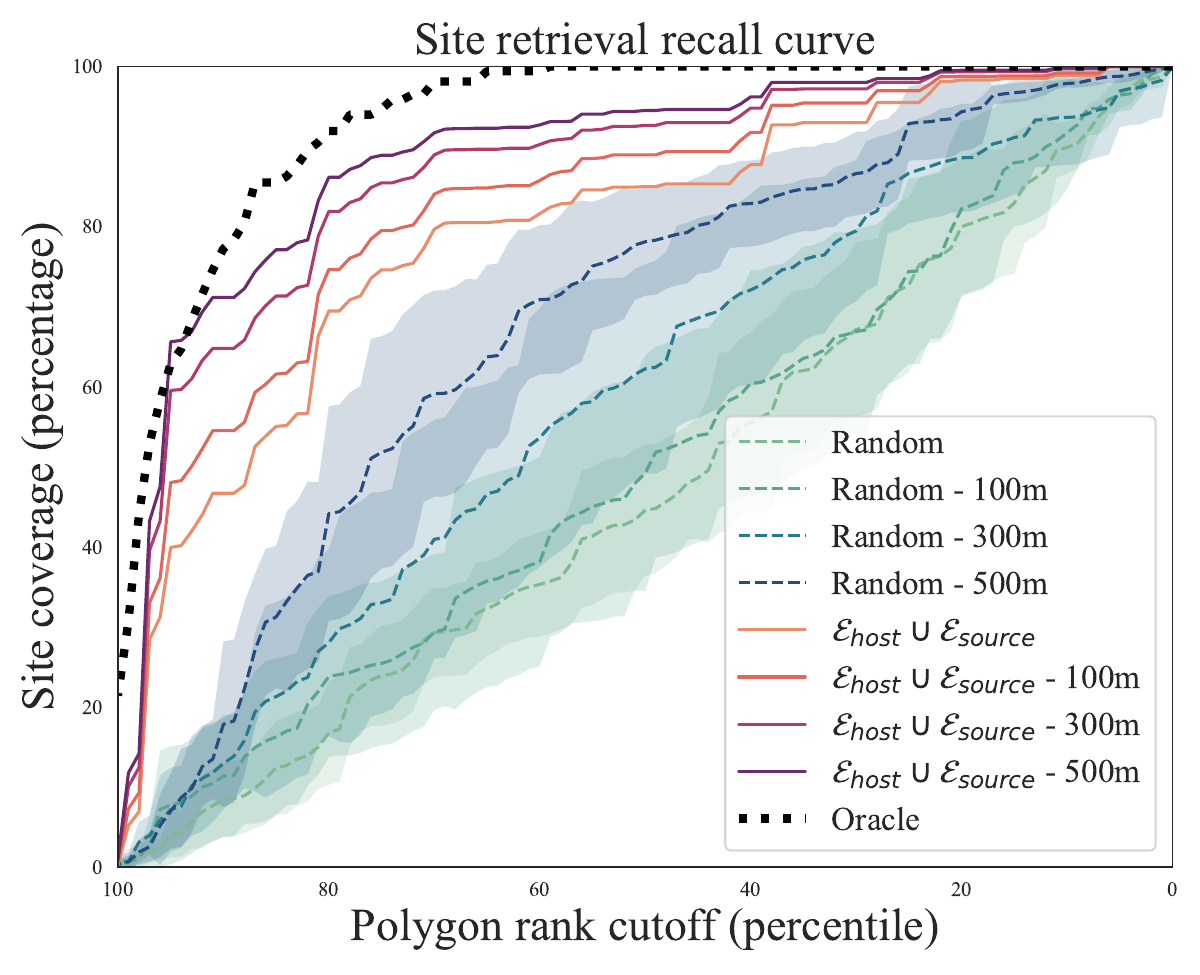}
    \caption{Recall curve of \queryplot, random ranking, and oracle ranking on site retrieval.}
    \label{fig:recall}
\end{figure}

\subsection{\queryplot as unsupervised predictive approach}
Since the evidence layers generated by \queryplot reflect how well each region matches with host and source rocks of tungsten skarn deposits, they can be used to approximate for mineral potential.
We treat this as a retrieval problem and quantitatively measure how well the polygons ranked by \queryplot align with known sites.
More specifically, we rank all polygons based on their similarity scores to the host query and source query, respectively. Then we use varying percentile values, e.g., from 1 percentile to 100 percentile at increment of 1, as cut-off threshold to filter high potential polygons. For each cut-off value, we count the number of known sites covered by high potential pologyons by either host or source queries and compute the recall value, i.e., percentage of covered sites in all known sites. 
The recall values and corresponding cut-off values are plotted in Fig.~\ref{fig:recall}.
As a reference, we also plot the recall curve for oracle ranking - ranking each polygon by the number of know sites it covers - and random ranking - randomly rank the polygon ten times and compute the mean and standard deviation of recall values.
For \queryplot and random ranking, we also buffered each high potential polygon by 100, 300, and 500 meters to check how much improvement can be made by extending their area.
From the plot we observe that:
(1) There is a steep increase on the recall curves of $\mathcal{E}_{host}\cup\mathcal{E}_{source}$ at the highest percentile cut-off values. Then, they gradually increase as the cut-off value is lower and plateaus toward the lower end of threshold values. This shows that polygons ranked higher by \queryplot does contain more known sites than those that are ranked lower, and those ranked at top $10\%$ are of high quality - they cover nearly $50\%$ of all the known sites.
(2) Random rankings produce near-diagonal recall curves. There is a significant gap between the \queryplot curve and the random ranking curve, even if the polygons were buffered by 500 meters.
(3) Buffering the high potential polygons ranked by \queryplot can significantly boost recall performance, e.g., polygons ranked at top $20\%$ ($80$ percentile) covered nearly $90\%$ of all the known sites. This indicates that many know sites are either directly covered by those high-ranking polygons or within a radius of $500$ meters to them.

These preliminary results show that \queryplot can retrieve tungsten skarn sites with decent recall, but it does not measure how precise these layers are.
Mineral site data is point-based, making it ineligible for computing precision metrics because area-based polygons contain an infinite number of points.
To overcome this issue, we perform a more extensive evaluation using another set of ground truth data from a previous tungsten mineral resource assessment conducted by human expert geologists~\cite{lederer2021tungsten}. In that study, the authors used various types of scientific data, including geological, geochemical, and geophysical maps as well as satellite imagery, to create permissive tracts and estimate undiscovered tungsten skarn deposits.
Because both the evidence layers and permissive tracts are area-based, it is more straightforward than using point-based mineral sites to compute evaluation metrics including precision measurements.
For the evidence layers, since each of them only reflect mineral prospectivity based on an individual characteristic, we compute their overlapping region using equations \ref{eq:buffer1}, \ref{eq:intersect}, and \ref{eq:buffer2}, incorporating the geological knowledge that skarns are most likely to form in the contact zone of \hostrocks and \sourcerocks.
We use the resulting \contact layer as \queryplot's prediction and compare it to the permissive tracts.

\begin{figure}[t]
  \centering
  \includegraphics[width=0.95\linewidth]{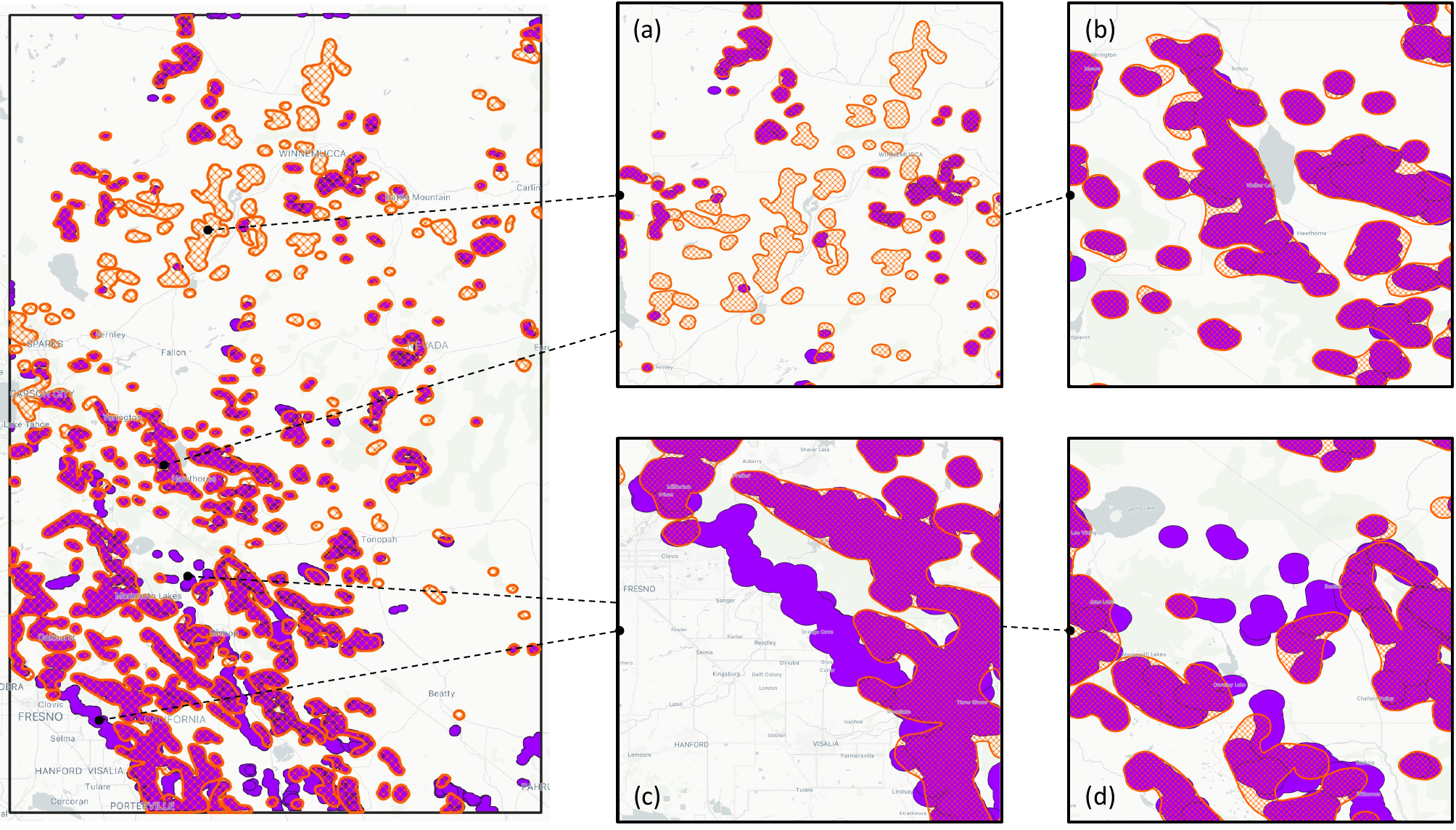}
  \caption{Comparison between \queryplot prediction and ground truth. Purple regions: \contact layer derived from high-ranking polygons in the \hostrocks and \sourcerocks evidence layers. Orange regions (mesh): permissive tracts from \cite{lederer2021tungsten}. (b)(c)(d): good alignment with ground truth with examples of `false positive' regions. (a): missed regions that are possibly due to lack of descriptive attributes in the query with keywords consistent with the deposit model.}
\label{fig:contact_permissive_tracts}
\end{figure}

The comparison is shown in Fig.~\ref{fig:contact_permissive_tracts}, where the purple regions indicate the \contact layer generated by \queryplot and the bright orange meshed regions are permissive tracts.
We observe that these two layers align well, with the major prospective regions in the southwest highlighted and many other small blobs across the study area predicted prospective. 
Some permissive tracts are missing in our predictions (Fig.~\ref{fig:contact_permissive_tracts}: (a)). A possible reason is that our approach only uses rock types information, while the ground truth permissive tracts were delineated by experts using rock types, age, and geophysical information.
We further examine the output by computing quantitative metrics. Let $\mathcal{E}^{pt}$ denote the permissive tracts layer evaluated by experts. We compute precision, recall, and F1 scores according to the following equations:


\begin{equation}
\label{eq:precision}
    \text{Precision}=\frac{\text{Area}(\mathcal{E}^+\cap\mathcal{E}^{pt})}{\text{Area}(\mathcal{E}^+)}
\end{equation}

\begin{equation}
\label{eq:recall}
    \text{Recall}=\frac{\text{Area}(\mathcal{E}^+\cap\mathcal{E}^{pt})}{\text{Area}(\mathcal{E}^{pt})}
\end{equation}

\begin{equation}
\label{eq:F1}
    \text{F1}=2\cdot\frac{\text{Precision}\cdot\text{Recall}}{\text{Precision}+\text{Recall}}
\end{equation}

\begin{figure}[t]
    \centering
    \includegraphics[width=0.5\linewidth]{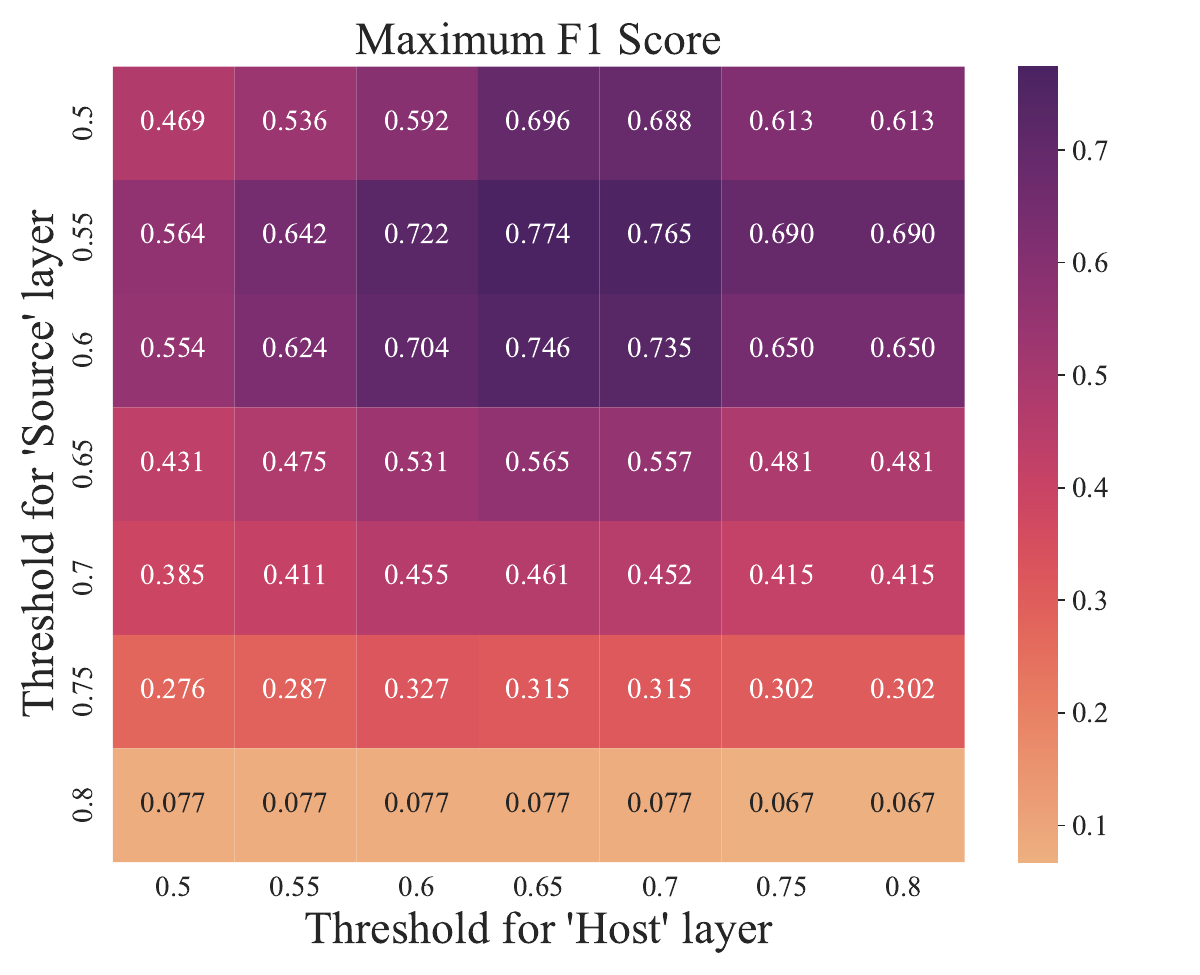}
    \caption{Best F1 scores with respect to different combination of thresholds for \hostrocks and \sourcerocks evidence layers.}
    \label{fig:heatmap}
\end{figure}
Note that the definitions of these area-based metrics differ from those typically used to evaluate machine learning classifiers, as we are comparing two different sets of geometric polygons and measuring how well one matches the another.
Since the thresholds for selecting top-ranked polygons and buffer distance values will affect the final output layer, and choosing these values often require domain knowledge and experience that is difficult to acquire under unsupervised setting, we did grid search over the hyperparameters to find the best configuration (a parameter sensitivity study is shown in Fig.~\ref{fig:heatmap}).
We used several different embedding models for this experiment, including the GTE series~\cite{zhang2024mgte} and BGE series~\cite{xiao2024c, chen2024m3}, both of which are general-purpose embedding models for measuring sentence-to-sentence similarity.
The results are summarized in Table~\ref{tab:metrics1}.
We observed that the evaluation metrics on the output layer can be drastically different depending on the embedding models used. The models' performance on mineral prospectivity mapping does not necessarily align with their performance on general embedding benchmarks. For example, the \texttt{large-v1.5} and \texttt{m3} versions of BGE~\cite{xiao2024c} embedding achieved higher average performance than the \texttt{base-v1.5} and \texttt{small-v1.5} versions over multiple tasks on Massive Text Embedding Benchmark (MTEB)~\cite{enevoldsen2025mmtebmassivemultilingualtext}, but they have lower intersection over union (IOU) and F1 scores in our tungsten skarn case study. We hypothesize that this is caused by the gap between these models' training data and the geologic map data used in this study. Unlike the common pattern that larger models will generally perform better on new tasks in the same domain, there is more uncertainty when applying these models in a different domain to measure semantic similarity between sentences that they have not been trained on during the pre-training stage. From this observation we also believe that embedding models for geoscience is urgently needed and will benefit many researchers in this community. We leave the training of such models as a topic to explore further in future works.

\begin{table}
\caption{\queryplot evaluation results on unsupervised Tungsten skarn MPM. All the numbers are best performance acquired through grid search over hyperparameters.}
\label{tab:metrics1}
  \begin{tabular}{cc|c|ccc}
    \toprule
    {\bfseries Approach} & {\bfseries embedding} & {\bfseries IOU} & {\bfseries Precision} & {\bfseries Recall} & {\bfseries F1}\\
    \midrule
    \multirow{2}{*}{\queryplot + GTE~\cite{zhang2024mgte}} & \texttt{base-v1.5} & 42.74 & 49.03 & 76.92 & 59.89 \\
    & \texttt{large-v1.5} & {\bfseries 63.14} & 79.89 & 75.07 & {\bfseries 77.40} \\
    \midrule
    \multirow{4}{*}{\queryplot + BGE~\cite{xiao2024c}} & \texttt{small-v1.5} & 60.63 & 68.45 & 84.14 & 75.49 \\
    & \texttt{base-v1.5} & 61.18 & 75.01 & 76.83 & 75.91 \\
    & \texttt{large-v1.5} & 48.53 & 51.97 & 87.99 & 65.34 \\
    & \texttt{m3}~\cite{chen2024m3} & 52.80 & 68.63 & 69.59 & 69.11 \\
    \bottomrule
  \end{tabular}
\end{table}

\subsection{\queryplot as additional input for supervised learning}
In the above evaluations, we generated evidence layers using only semantic information from geologic map data and descriptive deposit models, evaluated them by comparing directly with the permissive tracts created by experts, and showed both qualitatively and quantitatively that \queryplot is able to generate meaningful evidence map layers for mineral exploration.
However, this process is entirely unsupervised, as we do not use any existing tungsten skarn deposit site locations as labeled data, which have been shown to be valuable for training supervised machine learning models for predictive mineral mapping.
Since supervised approaches commonly take evidence map layers as input, it is straightforward to include evidence maps generated by \queryplot as additional layers. These additional layers could potentially provide richer information for classifiers to identify prospective locations.
To investigate this, we adopted the MPM workflow from~\cite{daruna2024gfm4mpm}, in which the authors trained classifiers using labeled mineral site locations to map input evidence layers into output prospectivity maps.
We selected 19 input layers based on geophysical (e.g., Gravity Bouguer, Isostatic Gravity, Magnetic RTP) and geochemical data (e.g., Potassium, Uranium, Thorium), and added one geology evidence layer, i.e., the \contact layer generated by \queryplot.
All layers were preprocessed using the pipeline proposed in~\cite{daruna2024gfm4mpm}, including rasterization, georeferencing, outlier removal, and normalization.
The processed layers were then used for self-supervised representation learning with an encoder-decoder model, followed by training an MLP classifier on known tungsten skarn site locations using the learned features. Performance was evaluated on a held-out set using metrics such as  Area Under the Precision-Recall Curve (AUPRC), Accuracy (ACC), Balanced Accuracy (Bal. Acc), Area Under the Receiver Operating Characteristic Curve (AUC), and Matthews Correlation Coefficient (MCC). The results are shown in Table~\ref{tab:metrics2}.
From these results, we observe that adding the additional geology evidence layer generated by \queryplot improves the MPM model performance across most metrics. This demonstrates that the evidence layer provides complementary geological information to the existing geophysical and geochemical layers, enabling the supervised classifier to more effectively associate input features with mineral potential.
The quality of evidence layers created by \queryplot depends on the underlying geological database. By using higher resolution, larger scale geologic map databases with more accurate descriptive attributes, \queryplot can improve the precision of evidence layers and thus provide reliable input for MPM.

\begin{table}
\caption{Evaluation results on supervised Tungsten skarn MPM. A geology evidence layer generated from \queryplot is used as an additional input channel to the binary classifier. \textcolor{lightgray}{\dag Metrics that are not suited for measuring performance on imbalanced data}.}
\label{tab:metrics2}
  \begin{tabular}{cc|cc}
    \toprule
    {\bfseries Data split} & {\bfseries Metric} & {\bfseries GFM~\cite{daruna2024gfm4mpm}} & {\bfseries GFM~\cite{daruna2024gfm4mpm}+\queryplot} \\
    \midrule
    \multirow{3}{*}{Validation} & AUPRC & {\bfseries 47.31} & 43.52 \\
     & AUC & 89.28 & {\bfseries 90.45} \\
     & F1 & 43.13 & {\bfseries 46.91} \\
    \hline
    \multirow{6}{*}{Test} & AUPRC & 41.53 & {\bfseries 43.04} \textcolor{green}{(+1.51)} \\
     & Bal. Acc & 72.39 & {\bfseries 78.37} \textcolor{green}{(+5.98)} \\
     & \textcolor{lightgray}{AUC\textsuperscript{\dag}} & \textcolor{lightgray}{91.85} & \textcolor{lightgray}{{\bfseries 92.91}} \\
     & \textcolor{lightgray}{ACC\textsuperscript{\dag}} & \textcolor{lightgray}{{\bfseries 95.31}} & \textcolor{lightgray}{94.28} \\
     & MCC & 46.57 & {\bfseries 48.26} \textcolor{green}{(+1.69)} \\
     & F1 & 48.97 & {\bfseries 50.40} \textcolor{green}{(+1.43)} \\
    \bottomrule
  \end{tabular}
\end{table}

\section{Prototype system}
\label{sec:prototype}
\begin{figure}[t]
  \centering
  \includegraphics[width=0.95\linewidth]{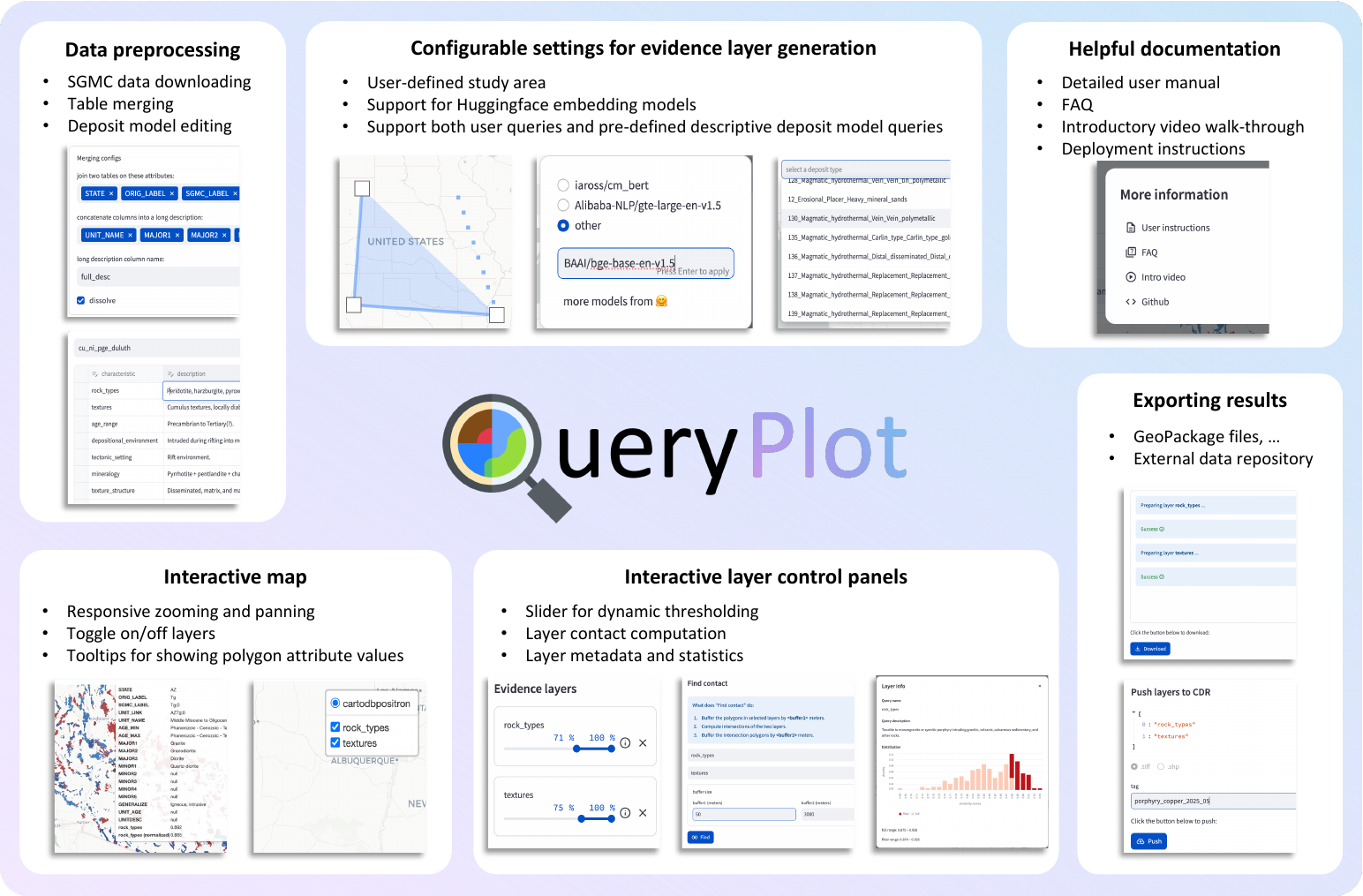}
  \caption{The \queryplot prototype system has a user friendly GUI that supports rich features, including flexible and customizable data pre-processing, interactive layer visualization, real-time filtering, and exporting processed layers to common formats readable by popular GIS software.}
\label{fig:prototype}
\end{figure}
To make \queryplot both usable and easily accessible to expert geologists who do not have extensive programming skills, we developed a prototype system that encapsulates the core functionality of \queryplot, i.e., evidence layer generation from geologic queries, as well as a suite of supporting tools, all within a web-based Graphical User Interface (GUI)(Fig.~\ref{fig:prototype}). The application is built using the Streamlit framework~\cite{streamlit} along with backend Python libraries including Transformers~\cite{wolf-etal-2020-transformers} and GeoPandas~\cite{jordahl_geopandas_2020}.
The entire system is containerized using Docker and can be deployed on popular cloud services such as Amazon Web Services (AWS).
In this section, we will highlight some of \queryplot's key features.

\subsection{Flexible and customizable data preparation}
Geologic map data processing is the critical first step in MPM. \queryplot provides several functionalities to simplify this process: (1) By default, \queryplot includes a preprocessed SGMC dataset that is ready to use out of the box. (2) Users can download the required files from the USGS official website directly onto the deployment server via a simple button click. Once downloaded, users can easily merge tables and preprocess the data with customizable settings, including selecting which attributes should be used to concatenate into long descriptions and whether to merge polygons. (3) Users can also upload their own shapefiles from their local machines to \queryplot.

MPM studies are often focused on a specific area rather than mapping mineral potential across the entire nation. \queryplot offers multiple ways to define the focus area for a study: (1) Users can draw a polygon interactively on a map and save it as a file for both the current and future studies. (2) Users can upload a shapefile containing their custom extent. (3) Users can select from a list of extent files hosted in an external data repository.
Because \queryplot uses the Transformers library, users benefit from the wide variety of embedding models available on HuggingFace~\cite{wolf2020transformers}. We include several embedding models as default options ( e.g., \texttt{BAAI/bge-base-en-v1.5}~\cite{xiao2024c}) for users who do not want to make a manual selection. Alternatively, experienced users can specify a preferred model name, and \queryplot will download the model weights from HuggingFace and use the chosen model for all subsequent operations.
These flexible options streamline the data preparation process for geologists, allowing them to focus on the important tasks of mineral exploration.

\subsection{Dual query mode: custom query and descriptive deposit models}
Once geologic map data are prepared, users can begin searching the database using natural language queries. \queryplot provides two approaches for searching polygons: (1) a custom mode, in which users can enter any textual query describing the properties they seek in the polygons, and (2) a deposit model mode, where users select from a set of characteristics pre-loaded from descriptive models of a certain deposit type they specify.
Both approaches use the same underlying embedding model, providing users the flexibility to generate evidence layers tailored to their specific needs.
The pre-loaded descriptive deposit models are also editable, enabling users to iteratively refine descriptions and track modifications for future MPM studies.

Compared to formal methods such as SQL queries, natural language querying in \queryplot enables semantic ``soft'' matching with similarity scores. This is particularly important in geoscience, where maps from different regions may include different terminology for the same materials and where geological terms evolve as new discoveries are made.
Formal queries must incorporate all possible term variants, leading to longer and harder-to-maintain statements. In contrast, the non-formal (natural language) approach is more adaptable to such shifts and produces search results in a more contextualized manner. 

\subsection{Interactive map visualization}
To facilitate investigation, \queryplot provides several operations for interacting with the generated layers: (1) An evidence layer control panel lists all currently active layers. Each layer is associated with a sub-panel, allowing users to dynamically adjust the minimum and maximum score thresholds and inspect layer metadata. As users adjust the threshold slider, the evidence layer map refreshes in real time, allowing geologists to observe relative potential rankings among polygons. The metadata window displays the original query and a distribution of similarity scores within the study area, allowing users to review the query and modify if necessary. 2) Tooltips are available for polygon attributes while users hover mouse on polygons, helping them to inspect whether polygon descriptions match with the intended query. 3) A layer switching panel allows users to toggle visibility of each layer, supporting focused inspection without distraction from other layers. 
\queryplot also integrates the \texttt{find contact} function, allowing users to derive new layers created from existing evidence layers. During this process, the buffering distance parameters $r_1$ and $r_2$ can be adjusted to achieve the desired output.
All these interactive features eliminate the time-consuming process of repeatedly exporting results and manually loading them into separate GIS software. Users can make adjustments within \queryplot and view the updated results in real-time, removing the need to switch between tools and thus further accelerating the mineral exploration workflow.

\section{Limitations and Opportunities}
\label{sec:limitation}
There are several limitations to this work:
(1) Artifacts in evidence maps: We observed artifacts—such as straight lines—appearing in some evidence maps. These arise because the SGMC is not a fully integrated geologic map database; geologic units have not been reconciled across state boundaries~\cite{horton2017state}. Such artifacts are undesirable, as geological properties do not conform to artificial state lines. Consequently, these issues may reduce the accuracy of generated geological evidence layers in reflecting the true potential for mineral deposits.
(2) Source geologic map data: \queryplot is limited by the spatial resolution and descriptive attributes of the source data. Incorporating more detailed maps with richer attribution can increase the precision of outputs, but this often comes at the cost of greater computation time and reduced areal coverage.
(3) Lack of domain-specific embedding fine-tuning: We utilized general-purpose embedding models without fine-tuning them on geological text corpora. Prior work in NLP has shown that fine-tuning embedding models on domain-specific data typically improves accuracy. However, this process requires large datasets of sentence pairs annotated with similarity scores, which are currently unavailable for geology. For this reason, we opted to use existing pre-trained models.
(4) Focus on data and algorithms, excluding economic and risk factors: Our present work focuses on data processing and algorithmic aspects of critical mineral exploration. Important factors such as economic viability and risk assessment are not considered and are beyond the scope of this study.

Based on the observations and findings from this work, we point out some directions worth further exploring in future research.
(1) Three-dimensional mineral potential mapping: Since minerals can occur at various depths underground, incorporating stratigraphic data to generate evidence layers across multiple depth intervals could significantly enhance \queryplot's utility for exploration geologists.
(2) Automated geodatabase comprehension and processing: Data collection and preprocessing represented a major time investment in our workflow; this is largely due to the heterogeneity of geological databases, which often feature diverse folder structures and file formats. Developing automated frameworks for data extraction, transformation, and loading (ETL) would substantially benefit future research in this area.

\section{Conclusion}
\label{sec:conclusion}
In this work, we introduced \queryplot, a system that leverages Natural Language Processing techniques for assisting geologists accelerate their mineral prospectivity mapping workflow.
\queryplot collects and processes geologic map data from the SGMC database, summarizes long academic documents into concise descriptive deposit models, and generates evidence layers by computing semantic similarity scores between polygon descriptions and user queries.
Through a case study on mapping prospective regions of Tungsten skarn deposits, we demonstrated that \queryplot can generate an evidence layer that is reasonably similar to human expert-generated permissive tracts.
Additionally, we showed that incorporating \queryplot layers can further boost supervised learning approach on its prospectivity mapping performance, as it provides complementary information to the other evidence layers originally used.
\queryplot's interactive web interface and flexible query modes make complex geoscientific tasks more accessible to domain experts, automating and accelerating their daily mineral exploration routines.
By open-sourcing our code and data, we hope to support further research in the AI for geoscience field and ultimately contribute to critical domains such as clean energy, digital infrastructure, and national security.




\begin{acks}
We gratefully thank Federico Solano and Jacob DeAngelo for their valuable feedback on an earlier version of this manuscript.
This work was funded by the Defense Advanced Research Projects Agency (DARPA) under Agreement No. HR00112390131.
Claims and conclusions of this article are of the authors and do not necessarily reflect those of DARPA.
\end{acks}

\bibliographystyle{ACM-Reference-Format}
\bibliography{sample-base}




\end{document}